\pdfoutput=1
\documentclass[a4paper,fleqn]{cas-dc}
\usepackage[numbers]{natbib}
\usepackage[utf8]{inputenc} 
\usepackage[T1]{fontenc}    
\usepackage{hyperref}       
\usepackage{url}            
\usepackage{booktabs}       
\usepackage{amsfonts}       
\usepackage{nicefrac}       
\usepackage{microtype}      
\usepackage{float}
\usepackage{appendix}
\usepackage{multirow}
\usepackage{lipsum}
\usepackage{indentfirst}
\usepackage{amsmath}
\usepackage{dsfont}
\usepackage{relsize}
\usepackage{svg}
\usepackage{amsthm}
\usepackage[]{graphicx}
\usepackage{graphicx}
\usepackage[english]{babel}
\usepackage{cases}
\usepackage{subfig}
\usepackage{caption}

\newcommand{\tool}{{CoreGen}\xspace}
\newcommand{\add}[1]{\textcolor{black}{#1}}
\newcommand{\nie}[1]{\textcolor{black}{#1}}

\def\tsc#1{\csdef{#1}{\textsc{\lowercase{#1}}\xspace}}
\tsc{WGM}
\tsc{QE}
\tsc{EP}
\tsc{PMS}
\tsc{BEC}
\tsc{DE}

\begin{document}
\let\WriteBookmarks\relax
\def\floatpagepagefraction{1}
\def\textpagefraction{.001}
\shorttitle{CoreGen: Contextualized Code Representation Learning for Commit Message Generation}
\shortauthors{Nie et~al.}

\title [mode = title]{CoreGen: Contextualized Code Representation Learning for Commit Message Generation}

\author[2]{Lun Yiu Nie}
\ead{lynie@link.cuhk.edu.hk}


\author[1]{Cuiyun Gao}
\cormark[1]
\ead{gaocuiyun@hit.edu.cn}

\author[3]{Zhicong Zhong}
\ead{zhongzhc3@mail2.sysu.edu.cn}


\author[2]{Wai Lam}
\ead{wlam@se.cuhk.edu.hk}

\author[4]{Yang Liu}
\ead{yangliu@ntu.edu.sg}

\author[1]{Zenglin Xu}
\ead{xuzenglin@hit.edu.cn}

\address[1]{Harbin Institute of Technology, Shenzhen, China}
\address[2]{The Chinese University of Hong Kong, Hong Kong, China}
\address[3]{Sun Yat-Sen University, Guangzhou, China}
\address[4]{Nanyang Technological University, Singapore}

\cortext[cor1]{Corresponding author}


\begin{abstract}
Automatic generation of high-quality commit messages for code commits can substantially facilitate software developers' works and coordination. However, the semantic gap between source code and natural language poses a major challenge for the task. Several studies have been proposed to alleviate the challenge but none explicitly involves code contextual information during commit message generation. Specifically,  existing research adopts static embedding for code tokens, which maps a token to the same vector regardless of its context. In this paper, we propose a novel \textbf{Co}ntextualized code \textbf{re}presentation learning strategy for commit message \textbf{Gen}eration (\tool). \tool first learns contextualized code representations which exploit the contextual information behind code commit sequences. The learned representations of code commits built upon Transformer are then fine-tuned for downstream commit message generation. Experiments on the benchmark dataset demonstrate the superior effectiveness of our model over the baseline models with at least 28.18\% improvement in terms of BLEU-4 score. Furthermore, we also highlight the future opportunities in training contextualized code representations on larger code corpus as a solution to low-resource tasks and adapting the contextualized code representation framework to other code-to-text generation tasks.

\end{abstract}




\maketitle
\section{Introduction}
\begin{figure}
    \vspace{0.15cm}
     \centering
     \includegraphics[width=0.48\textwidth]{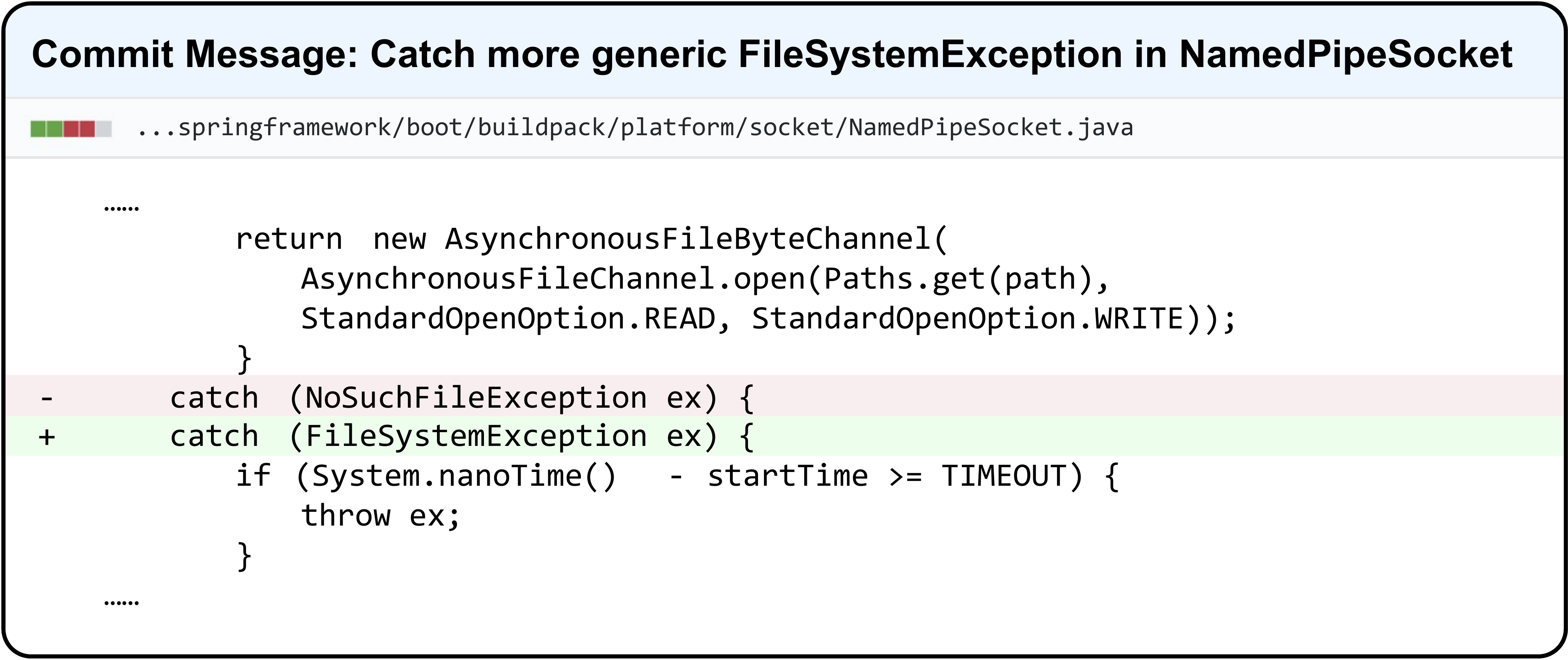}
     \caption{An example of code commit and its corresponding commit message.} 
     \label{fig:example}
\end{figure}

\add{Massive amount of source code are being produced in people's daily lives and works, thus bridging the gap between source code and natural language has become a practically useful but challenging task.} Mitigating such gap will enable the semantics of source code being connected to natural language, which is critical for solving many important tasks, such as commit message generation. In the life cycle of software development, the commit messages on version control systems (e.g., GitHub, GitLab) are essential for developers to document the abstract code changes in high-level natural language summaries. One example of code commit message is shown in Figure \ref{fig:example}, where a line of code has been updated for more generic exception handling. The line marked with ``+'' in green background is the newly added code while the line in red background marked with ``-'' indicates code been deleted, and the corresponding commit message is shown at the top. High-quality commit messages allow developers to comprehend the high-level intuition behind the software evolution without diving into the low-level implementation details, which can significantly ease the collaboration and maintenance of large-scale projects \cite{buse2010automatically}.


\add{In practice, however, the quality of commit messages is not guaranteed.} Dyer et al. \cite{dyer2013boa} report in their study that around 14\% of the Java projects on SourceForge leave commit messages completely blank. \add{Developers' intentional or unintentional negligence due to their lack of time and motivation both result in the sacrifice of commit messages' quality, let alone writing meaningful yet concise commit messages requires developers to grasp the essential ideas behind the code changes and explicitly summarize them from a holistic perspective, which is a skill that relies heavily on individual developer's expertise.} Even for the experienced experts, writing high-quality summaries for massive code commits still poses considerably extra workload. 


Therefore, automatic generation of high-quality commit messages becomes necessitated and many approaches have been proposed to address the needs. At the earlier stage, researchers adopt pre-defined templates to generate commit messages from extracted information \cite{buse2010automatically,cortes2014automatically,linares2015changescribe,shen2016automatic}. However, these rule-based methods require human developers to manually define templates. For the code commits that do not match any of the pre-defined rules, their approaches may fail in generating meaningful commit messages. For example, in Shen et al.'s work \cite{shen2016automatic}, their defined rules can only handle four stereotypical types of code commits \add{straightforwardly} as filling in the template ``Add [added information] at [method name]'' for in-method sentence modifications. To solve this issue, later works \cite{huang2017mining, liu2018nngen} leverage information retrieval techniques to reuse existing commit messages for incoming code commits. In spite of the improved flexibility, the quality of retrieved messages is still constrained by inconsistent variable/function names.

With the advancement of neural machine translation (NMT), recent researchers treat commit message generation as a code-to-text translation task and utilize deep neural networks to model the relationship between code commits and commit messages \cite{jiang2017automatically, loyola2017neural,xu2019commit,liu2019generating}, which are claimed to achieve the state-of-the-art performance on the benchmark. 

Despite the comparative successes of deep learning models in code commit message generation, all of these studies suffer from three critical limitations. First, existing research generally adopts static embedding methods for code representation, mapping a code token to an identical vector \add{representation} regardless of its context. However, code data are essentially different from textual data considering the semantic gap between source code and natural language. For example, a single token alone, if in textual data, can represent partial semantics, but usually cannot convey any meaningful information in source code without a context. Second, prior studies simply take the whole code commit snippet as input without attending explicitly to the changed fragments. Third, existing NMT models for commit message generation are all recurrent-based, which has been evidenced to suffer from long-term dependency issue \cite{DBLP:journals/corr/BahdanauCB14}.

\begin{figure*}
     \centering
     \includegraphics[width=\textwidth]{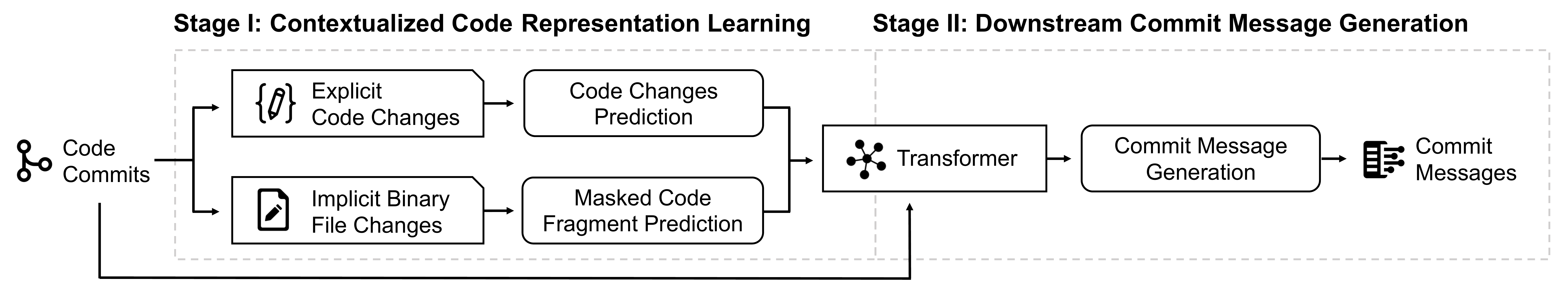}
     \caption{Workflow of \tool's two-stage framework.
     } 
     \label{fig:flow}
\end{figure*}

In this paper, we propose a novel two-stage framework for code commit message generation, named \tool, to address the above limitations. Inspired by the recent success of pre-trained language models \cite{peters2018deep, devlin2018bert, radford2019language, song2019mass}, we propose to model the code semantics with contextualized code representations, endowing one identical code token with different embeddings based on the respective contextual information. By training the model to predict code changes, the model is also guided to put more attention on the changed fragments rather than the whole commit snippets. At the second stage, the learned code representations are preserved, and further fine-tuned for downstream commit message generation. Both stages are implemented based on Transformer \add{to overcome the drawbacks of recurrent-based models}. Experimental results on benchmark dataset indicate that \tool achieves the new state-of-the-art on code commit message generation. 

The main contributions of our work are summarized as follows: 
\begin{itemize}
   \item We propose a two-stage framework named \tool that \add{first in the field highlights the divergence between the two categories of code commits, and effectively exploits contextualized code representations by predicting either the code changes or masked code fragment according to the nature of commits, which is built upon the Transformer model, yet can be easily adapted to other model architectures such as RNN. }
   
   \item We empirically show that \tool significantly outperforms previous state-of-the-art models with at least 28.18\% improvement on BLEU-4 score. Our in-depth studies and comparison experiments \add{further demonstrate \tool's superior usefulness in speeding up the model convergence and performing well under low-resource settings.}
   
    \item \add{We highlight \tool's potentials in generalizing to other low-resource tasks by adopting similar contextualized representation learning tasks, and a promising future research direction of improving \tool by modeling more complicated code structural information. We have released our implementation details publicly\footnote{https://github.com/Flitternie/CoreGen} to facilitate future research. }
\end{itemize} 

The rest of the paper is structured as follows. Section \ref{sec:approach} introduces our proposed two-stage framework. Section \ref{sec:experiment} and Section \ref{sec:result} describe the experimental setups and results. Section \ref{sec:discussion} provides \add{some detailed discussion around \tool}. Finally, Section \ref{sec:literature} reviews the related works and Section \ref{sec:con} concludes the paper.
\begin{figure*}
     \centering
     \includegraphics[width=\textwidth]{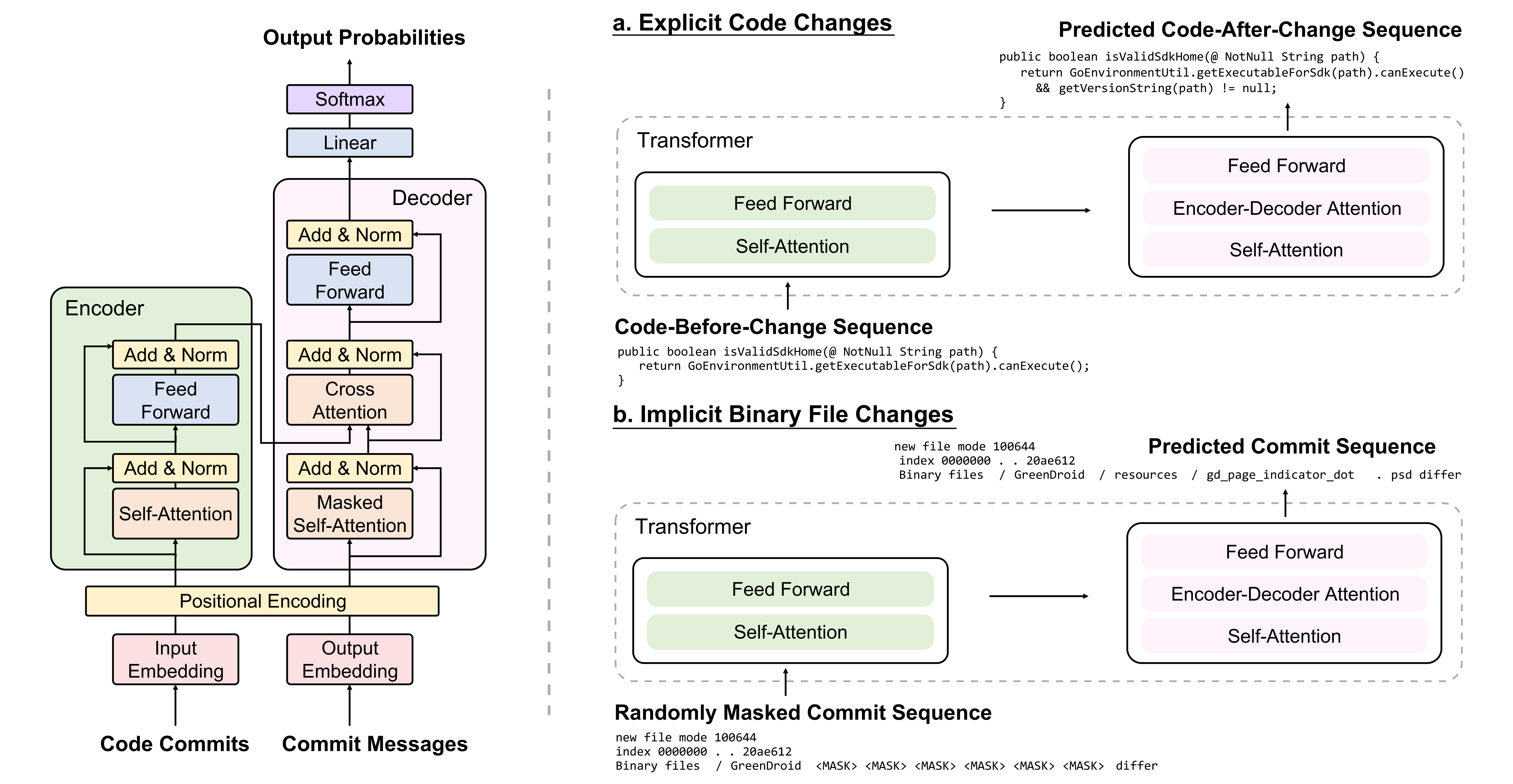}
         \caption{Overall model architecture. Left part represents the standard architecture of Transformer \cite{vaswani2017attention}. Right part describes the two contextualized code representation learning tasks proposed for the two respective categories of code commits.} 
     \label{fig:model}
\end{figure*}

\section{Approach}\label{sec:approach}
In this section, we introduce our approach, \underline{Co}ntextualized Code \underline{Re}presentation Learning for Commit Message \underline{Gen}eration (\tool), a two-stage framework for commit message generation. An overview of \tool is shown in Figure \ref{fig:flow}. \tool first learns contextualized code representation for the two separate categories of code commits via their respective representation learning strategy at Stage I, as illustrated in the right part of Figure \ref{fig:model}, then fine-tunes the whole model for downstream commit message generation task at Stage II, as shown in the left of the same figure. 

\add{Unlike previous works that neglect the divergence between the two categories of code commits, we recognize such difference and deliberately propose separate representation learning strategies to achieve more effective exploitation of the code contextual information. Also, please note that \tool's framework is orthogonal to the selection of specific model architecture, and can be easily generalized to include other code representation learning tasks, as explained in details in Section \ref{sec:discussion}. }


\begin{figure}
\centering
\subfloat[A code commit with explicit code changes]{\includegraphics[width=0.48\textwidth]{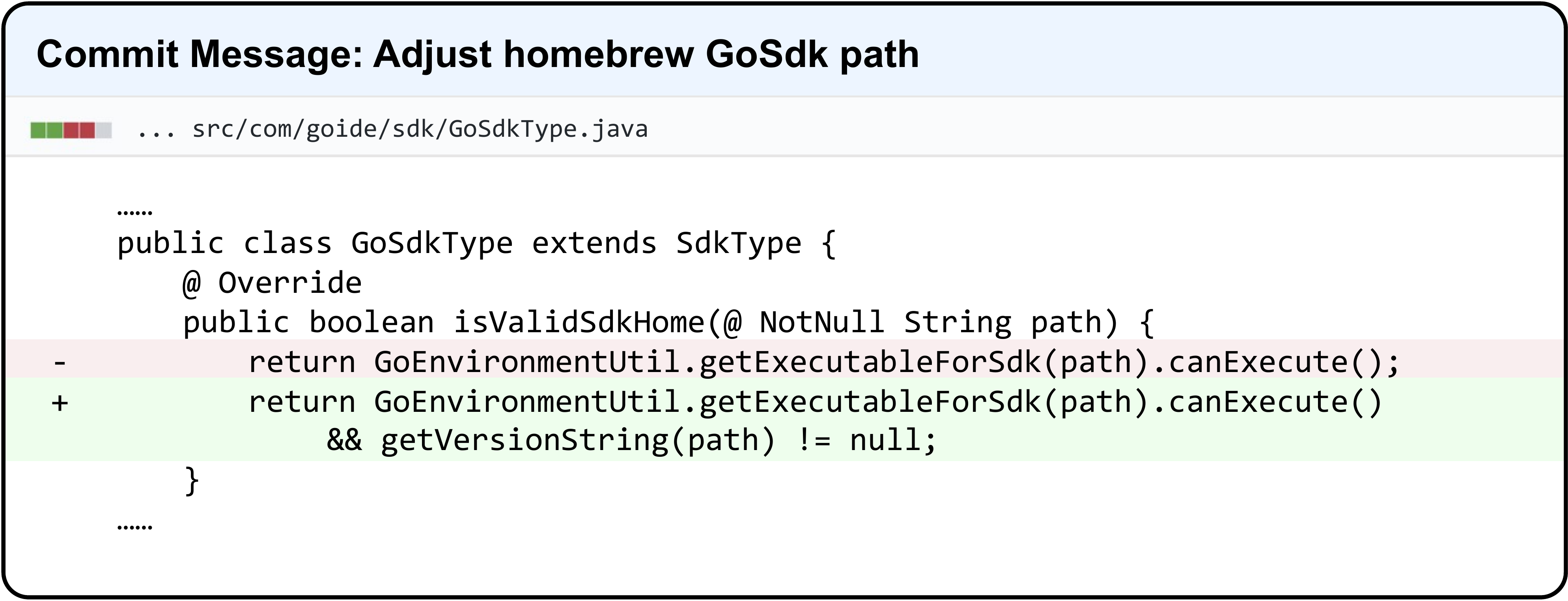}}\hfill
\subfloat[A code commit with implicit binary file changes]{\includegraphics[width=0.48\textwidth]{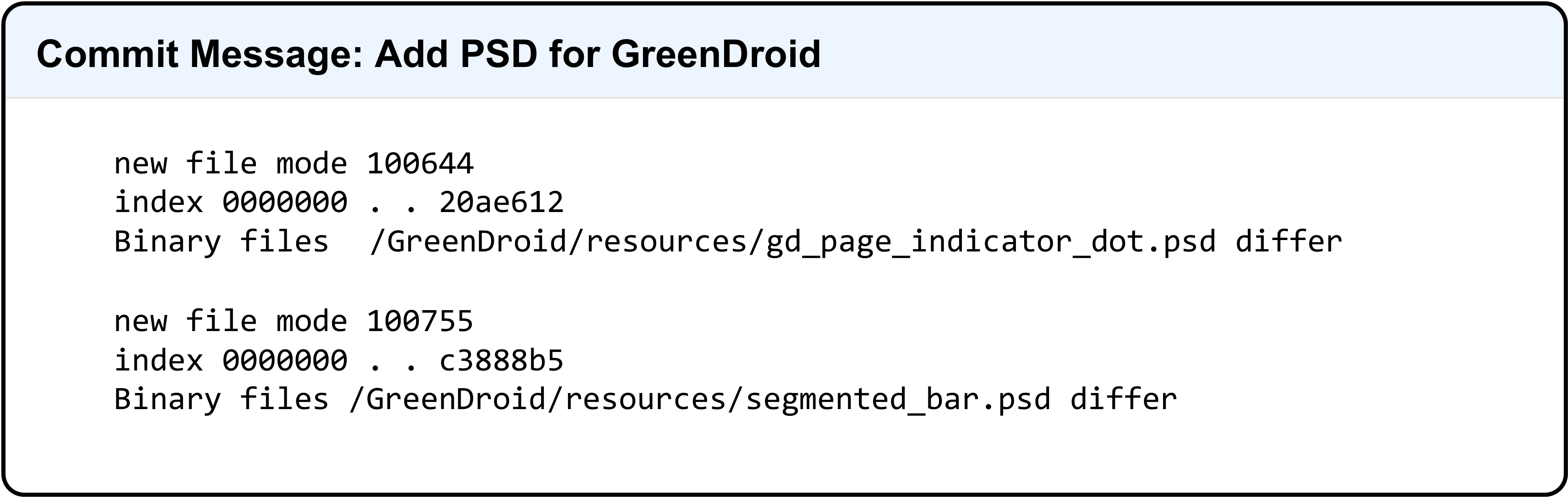}}
\caption{Examples of two code commit categories: \add{(a) explicit code changes, in which line-by-line code modification can be easily detected; and (b) implicit binary file changes, where content changes cannot be examined in details.} }
\label{fig:types}
\end{figure}

\subsection{Stage I: Contextualized Code Representation Learning}
Code commits can be naturally categorized into two types: one with explicit code changes and another with implicit binary file changes, by their respective features as illustrated in Figure \ref{fig:types}. \add{To enrich code representations with the contextual information for more accurate commit message generation,} for each code commit, \tool performs automatic categorization, then trains the Transformer via its corresponding representation learning task to exploit contextualized code representations. The details are elaborated as the following. 

\subsubsection{Code Changes Prediction}
The first category of code commits includes explicit code changes such as line addition, deletion, or modification. Generally, the lines are marked with special tokens at the beginning, e.g., ``+'' for addition and ``-'' for deletion. These changed code statements, comparing to the unchanged part of source code, play a much more crucial role in code commit message generation, since commit messages, by definition, should be summarizing the changes instead of the whole code snippets. For example, in Figure \ref{fig:types}(a), the commit message is primarily describing the changed code fragments (i.e., the lines in colored background) rather than the whole snippet that implements the class methods. Therefore, code changes prediction is designated as the contextualized code representation learning task for this category of code commits.

Given a code commit sequence $X$, we preprocess and split the source code sequence into code-before-change and code-after-change subsequences, denoted as $X^{\textit{before}}$ and $X^{\textit{after}}$ respectively, by locating the special tokens marked in the code commits. If explicit code changes are identified, we train the Transformer network to predict the changes by modeling the relationship between $X^{\textit{before}}$ and $X^{\textit{after}}$. Transformer \cite{vaswani2017attention} is a self-attention-based encoder-decoder architecture that has achieved the state-of-the-art performance in many machine translation benchmarks. In general, the encoder module reads the input sequence as a sequence of hidden representations and the decoder module converts the hidden representations into an output sequence by generating one token at a time. Specifically, we feed the code-before-changes sequences into the Transformer as input to predict the corresponding code-after-change sequences, as illustrated in the top right part of Figure \ref{fig:model}. Log likelihood is used as the objective function:
\begin{equation}
    \mathlarger{\mathcal{L}}_{a}=-
    \sum_{X\in \mathds{C}_1} \sum_{i}\log \mathcal{P}(x^{\textit{after}}_i|X^{\textit{before}};\theta)\ ,
\end{equation}
where $x^{\textit{after}}_i$ represents the $i$-th code token in the code-after-changes sequence $X$ to be predicted, $\mathds{C}_1$ refers to the code commit subcorpora with explicit code changes, i.e., $X^{\textit{after}} \neq X^{\textit{before}}$, and $\theta$ represents the Transformer model parameters to be learned. 

By predicting the code changes from their respective contexts, we explicitly guide the Transformer to put more attention to the changed code fragments and build up connections between the contextual code tokens and changed code tokens, thereby enriching the representations of code changes with their contextual information. 

\subsubsection{Masked Code Fragment Prediction}
Another category of commits include implicit binary file changes where detailed modifications inside the binary files are not visible. For example, in Figure \ref{fig:types}(b), two binary files are added in the commit while no content changes can be examined in detail. To model the context of file changes, we randomly mask a fragment of the code commit sequence and learn the contextualized code representations by predicting the masked tokens from the remaining ones.

Instead of randomly masking only one token as in BERT \cite{devlin2018bert}, we mask a fragment of tokens for the Transformer to model the context, considering that one single token in code snippet is generally of limited semantics. Given a code commit sequence $X$, we split it into $n$ different lines $\{X^1, X^2, ..., X^n\}$ using the special token ``<nl>'' and randomly mask a certain fragment of the longest line denoted as $X^k$. Then we train the Transformer to predict the masked code fragment $X^k_{u:v}$ from its context $\{X^1, ..., X^k_{\setminus u:v}, ..., X^n\}$, as illustrated in the bottom right part of Figure \ref{fig:model}. Log likelihood is again used as the objective function:
\begin{equation}
    \mathlarger{\mathcal{L}}_{b}=-
    \sum_{X\in \mathds{C}_2}\sum_{i=u}^{v} \log \mathcal{P}(x^k_{i}|X^1, ..., X^k_{\setminus u:v}, ..., X^n;\theta)\ ,
\end{equation}
where $x^k_i$ represents the $i$-th token in the masked line $X^k$ to be predicted, $\mathds{C}_2$ refers to the code commit subcorpora with implicit file changes, i.e., $X^{\textit{after}} = X^{\textit{before}}$, and the mask length is determined together by a mask rate $\phi$ and the length of the longest line $|X^k|$:
\begin{equation}
    |X^k_{u:v}|=\phi \cdot |X^k|\ .
\end{equation}
By predicting the masked code fragments based on their contexts, contextual information is incorporated into Transformer's embedding layer and encoder-decoder modules, which altogether produce contextualized code representations. \\

Finally, the overall objective of the first stage's training can be expressed as:  
\begin{equation}
    \mathlarger{\mathcal{L}}_{\text{I}}(\theta;\mathds{C}) =\frac{1}{|\mathds{C}|}(\mathcal{L}_a+\mathcal{L}_b)\ ,
\end{equation}
where $\mathds{C}$ refers to the entire training corpus that consists of \add{two categories of subcorpora} $\mathds{C}_1$ and $\mathds{C}_2$. As this stage ends, the learned contextualized representations of code commits are then transferred to Stage II for further fine-tuning. 

\subsection{Stage II: Downstream Commit Message Generation}
At Stage II, we transfer the contextualized code representations along with the Transformer model parameters (i.e., $\theta$) learned from Stage I for downstream commit message generation training. The whole Transformer network is optimized throughout the fine-tuning process with back-propagation applied to all layers. Specifically, given a code commit sequence $X$, the model is fine-tuned to predict its corresponding commit message sequence $Y$ with the following objective function:
\begin{equation}
    \mathlarger{\mathcal{L}}_{\text{II}}(\theta;\mathds{C})=-\frac{1}{|\mathds{C}|}\sum_{X\in \mathds{C}} \sum_{i} \log \mathcal{P}(y_i|X;\theta)\ ,
\end{equation}
where $y_i$ represents the $i$-th commit message token to be generated, $\mathds{C}$ refers to the same training corpus as in Equation (4) and $\theta$ represents the model parameters that have been trained in Stage I. To ensure a complete parameter migration with all the contextual information maintained, \add{the model architecture in Stage II is kept consistent to Stage I. }

\section{Experimental Setup}\label{sec:experiment}
In this section, we describe the benchmark dataset, metrics, baseline models, and parameter settings used in our evaluation.

\subsection{Dataset}
We conduct evaluation experiments based on the benchmark dataset released by Liu et al. \cite{liu2018nngen}, which is a cleansed subset of Jiang et al.'s published dataset \cite{jiang2017automatically}. The original dataset contains $\sim$2M pairs of code commits and corresponding commit messages collected from popular Java projects in GitHub. Liu et al. further cleanse the dataset by tokenizing the code commit sequences with white space and punctuation, removing non-informative tokens (e.g., issue IDs and commit IDs), and filtering out the poorly-written commit messages \cite{liu2018nngen}. This leaves us $\sim$27k pairs of code commit and commit message, which \add{have been} split into training set, validation set and test set at an approximate ratio of 8:1:1.


\begin{table*}
 \centering
\caption{Comparison of \tool with the baseline models using different evaluation metrics.}
\label{result}
\scalebox{1.0}{
\begin{tabular}{cc|c|c|c|c|c}
\toprule 
\multicolumn{2}{c|}{\multirow{1}{*}{\textbf{Model}}}                                    &  \multirow{1}{*}{\textbf{BLEU-4}}    & \multirow{1}{*}{\textbf{ROUGE-1}}           & 
\multirow{1}{*}{\textbf{ROUGE-2}}           & 
\multirow{1}{*}{\textbf{ROUGE-L}}               & \multirow{1}{*}{\textbf{METEOR}} \\ 
\midrule
\multicolumn{1}{c|}{\multirow{3}{*}{\textbf{Baselines}}} & \multicolumn{1}{l|}{NMT} 
&  14.17 & 21.29 &  12.19 & 20.85  & 12.99 
\\\multicolumn{1}{c|}{} & \multicolumn{1}{l|}{NNGen}
&  16.43  &  25.86 & 15.52  & 24.46 &  14.03
\\\multicolumn{1}{c|}{} & \multicolumn{1}{l|}{PtrGNCMsg} 
& 9.78  &  23.66  &  9.61 & 23.67   & 11.41
\\ 
\midrule
\multicolumn{1}{c|}{\multirow{2}{*}{\textbf{Ours}}}  
& \multicolumn{1}{l|}{\tool\textsubscript{II}}
&  18.74 & 30.65 & 18.06 & 28.86 & 15.18
\\ \multicolumn{1}{c|}{} &  \multicolumn{1}{l|}{\tool} & \textbf{21.06} & \textbf{32.87} & \textbf{20.17} &  \textbf{30.85} & \textbf{16.53}
\\
\bottomrule
\end{tabular}}
\end{table*}

\subsection{Evaluation Metrics}
We verify the effectiveness of \tool with automatic evaluation metrics that are widely used in natural language generation tasks, including BLEU-4, ROUGE and METEOR. BLEU-4 measures the 4-gram precision of a candidate to the reference while penalizes overly short sentences \cite{papineni2002bleu}. BLEU-4 is usually calculated at the corpus-level, which is demonstrated to be more correlated with human judgments than other evaluation metrics \cite{DBLP:conf/emnlp/LiuLSNCP16}. Thus, we use corpus-level BLEU-4 as one of our evaluation metrics.

To mitigate BLUE-4's preference on long-length commit messages, we also employ ROUGE, a recall-oriented metric particularly proposed for summarization tasks, to evaluate the quality of generated commit messages \cite{lin2002rouge}. In this paper, we compute the ROUGE scores on unigram (ROUGE-1), bigram (ROUGE-2) and longest common subsequence (ROUGE-L) respectively. Taking advantages of the weighted F-score computation and penalty function on misordered tokens, METEOR is \add{another} natural language generation metric used in our experiments \cite{lavie2007meteor}.

\subsection{Baseline Models}
We compare the proposed \tool with the following baseline models in the experiments. For the sake of fairness, we apply the same cleansed benchmark dataset for evaluating the baseline models and \tool. 

\begin{itemize}
    \item \textbf{NMT.} NMT model uses an attentional RNN Encoder-Decoder architecture to translate code commits into commit messages \cite{loyola2017neural, jiang2017automatically}. Specifically, Jiang et al. \cite{jiang2017automatically} implement the NMT model using a TensorFlow built-in toolkit named Nematus \cite{sennrich2017nematus}. 
    
    \item \textbf{NNGen.} NNGen is a retrieval-based model that leverages nearest neighbor algorithm to reuse existing commit messages \cite{liu2018nngen}. It represents each code commit sequence as a ``bags of words'' vector and then calculates the cosine similarity distance to retrieve top $k$ code commits from the database. The commit with the highest BLEU-4 score to the incoming commit is thereafter regarded as the nearest neighbor and the corresponding commit message is then output as the final result. 
    
    \item \textbf{PtrGNCMsg.} PtrGNCMsg \cite{liu2019generating} is another RNN-based Encoder-Decoder model that adopts a pointer-generator network to deal with out-of-vocabulary (OOV) issue. At each prediction time step, the RNN decoder learns to either copy an existing token from the source sequence or generate a word from the fixed vocabulary, enabling the prediction of context-specific OOV tokens in commit message generation. 
\end{itemize}

\subsection{Parameter Setting}
We conducted experiments on different combinations of hyperparameters to optimize \add{\tool's end-to-end performance of on the validation set of the benchmark}. Specifically, we feed the code commits and commit messages into \tool with a shared vocabulary of 55,732 unique tokens. The input dimension of the tokens is set as 512. The input embeddings of the code tokens are randomly initialized at the beginning of Stage I, then get trained \add{throughout the contextualized code representation learning procedure. The learned embeddings are next transferred} for Stage II's downstream commit message generation, during which the code embeddings are further fine-tuned to be task-aware. For the Transformer, both the encoder and decoder modules are composed of 2 identical layers while each layer includes 6 parallel multi-attention heads.

For training, we use Adam optimizer \cite{kingma2014adam} with batch size equals to 64 and the learning rate is adjusted dynamically in line with the original implementation with the warm-up step set to 4000 \cite{vaswani2017attention}. The mask rate $\phi$ for Stage I's \textit{Masked Code Fragment Prediction} is set to 0.5. \add{All these hyperparameter settings are tuned on the validation set.} A detailed analysis about the impact of hyperparameters on \tool's performance can be found in Section \ref{sec:setting}.

\section{Experimental Results}\label{sec:result}
\subsection{Result Analysis}\label{sec:main_result}
Table \ref{result} shows the experimental results of our model and the baselines. \tool outperforms baseline models across all evaluation metrics with at least 28.18\%, 26.12\% and 17.82\% improvement on BLEU-4, ROUGE-L and METEOR scores, respectively. We attribute this to its effectiveness for attending to the critical segments of code snippets, i.e., the changed code fragments. \add{Besides, comparing to PtrGNCMsg that employs an extra pointer-generator network to copy the context-specific OOV tokens, \tool's superior performance further supports our claim that exploiting the code contextual information can achieve a more accurate modeling of the context-specific tokens (e.g., variable/function names), leading to an elegant solution to the OOV issue.}

\begin{figure}
     \centering
     \includegraphics[width=.45\textwidth]{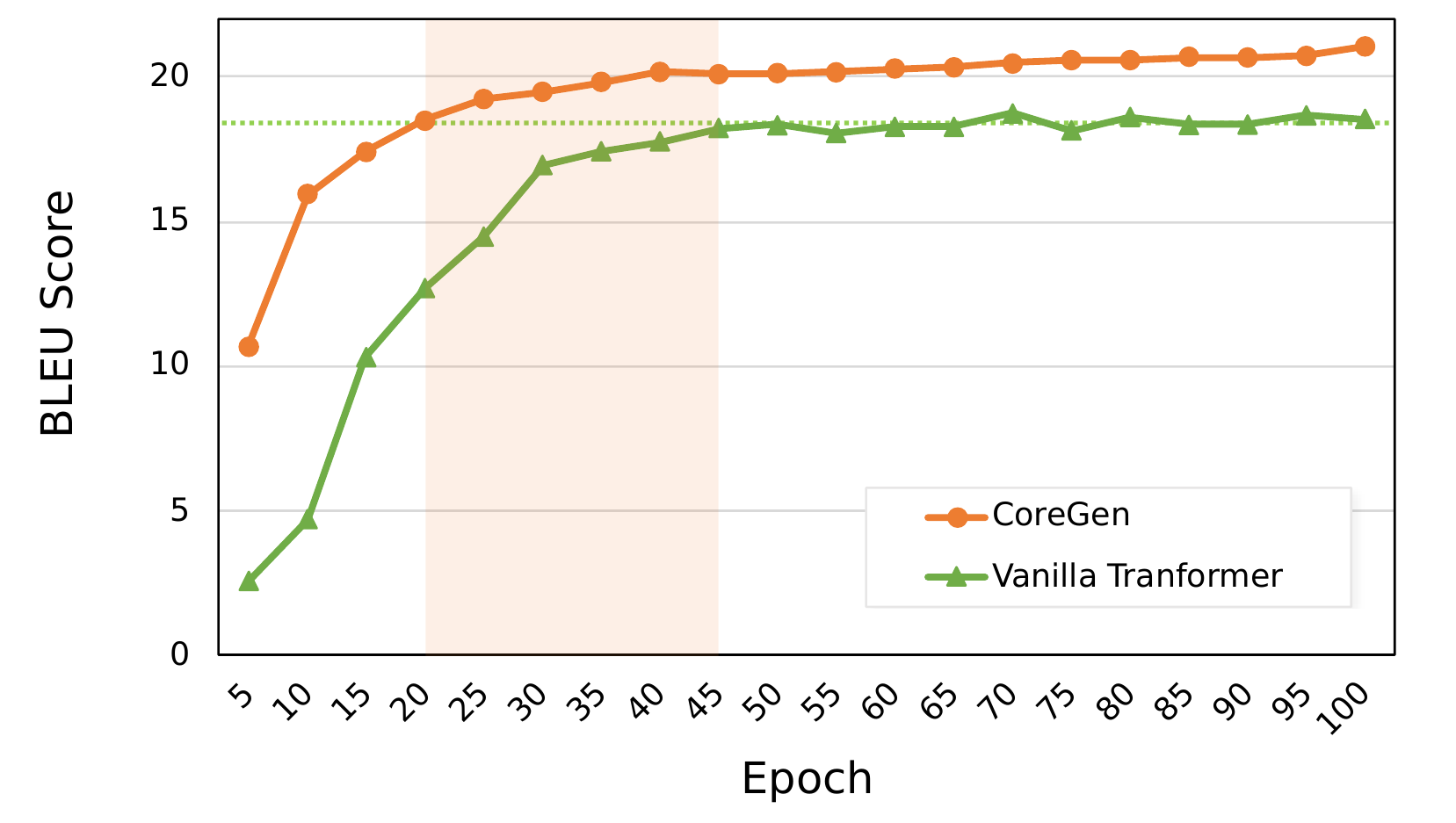}
         \caption{Convergence between Transformer and \tool. }
     \label{fig:conv}
\end{figure}

Besides, \tool's contextualized code representation learning procedure can also speed up model's convergence. In our experiment comparing vanilla Transformer and \tool on model's convergence along the Stage II training procedure, as Figure \ref{fig:conv} illustrates, \tool can converge faster to achieve equivalent generation quality as the vanilla Transformer model at 25 training epochs ahead.

\begin{figure}
     \centering
     \includegraphics[width=.35\textwidth]{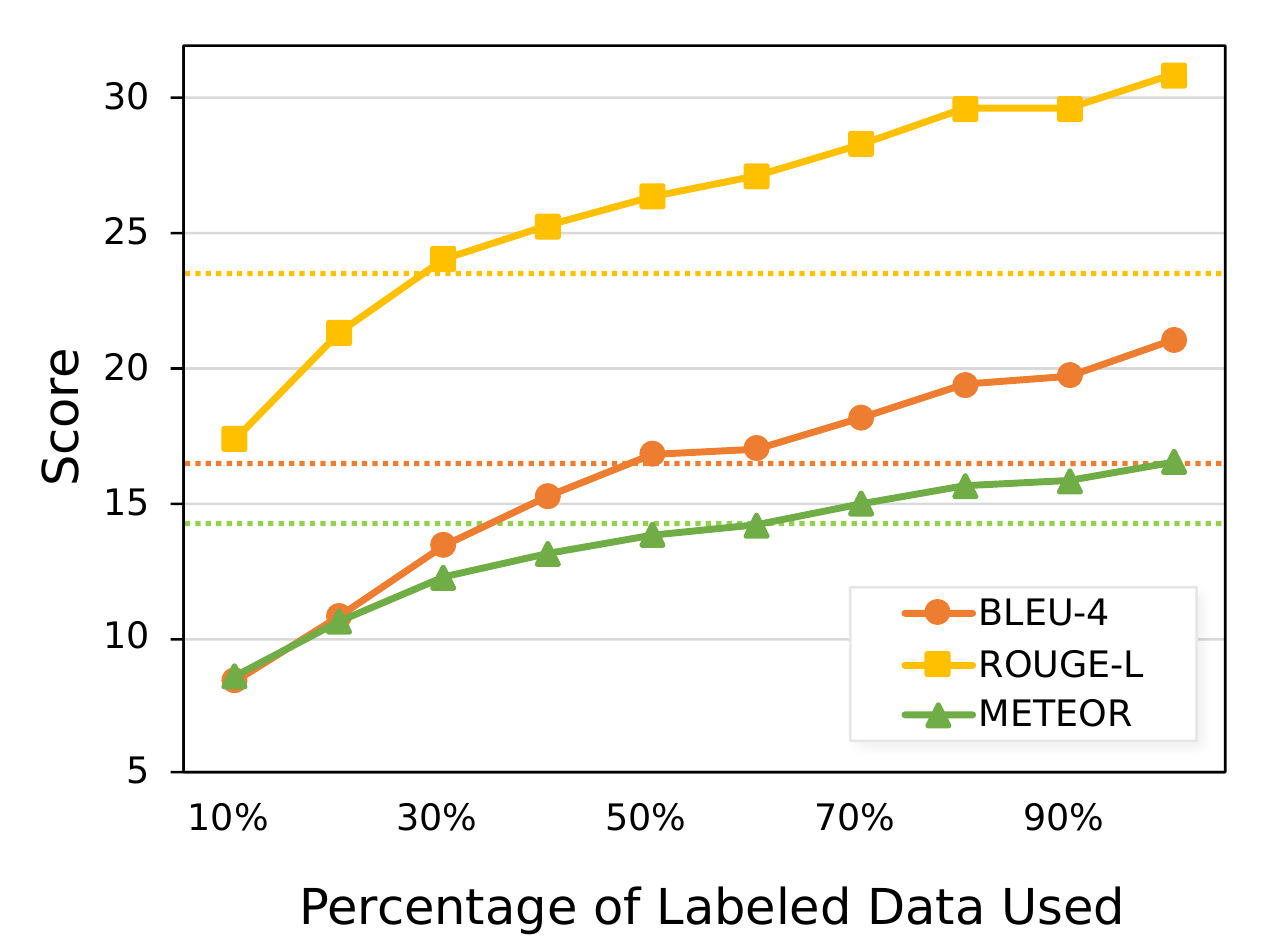}
         \caption{\tool's performance under low-resource settings. Dotted horizontal lines indicate the best performance achieved by baselines.} 
     \label{fig:lowres}
\end{figure}

\add{In practice, collecting high-quality commit messages is difficult since substantial efforts are required for differentiating messages' quality \cite{liu2018nngen}. Therefore, to simulate real-life usage, we further validate \tool's generalization ability under low-resource settings. After using the whole training corpus for contextualized code representation learning, we adjust the amount of labels (i.e., commit messages) available for Stage II's supervised fine-tuning. As shown in Figure \ref{fig:lowres}, \tool outperforms the baseline models (annotated as the dotted lines) by making use of only 50\% of the labels. This inspiring result not only indicates the strong generalization ability of our proposed contextualized code representation learning strategies, but also suggests promising future research directions, such as training the contextualized code representation on larger corpus as a general solution to code-related tasks, especially when under the low-resource settings. }


\subsection{Ablation Study}
To further validate the usefulness of contextualized code representation learning, we also compare \tool with an ablated method \tool\textsubscript{II} that performs Stage II's downstream fine-tuning from scratch on Transformer and skips Stage I's representation learning procedure. 

\begin{figure*}
\centering
\subfloat[Impact of mask rate]{\includegraphics[width=0.3\textwidth]{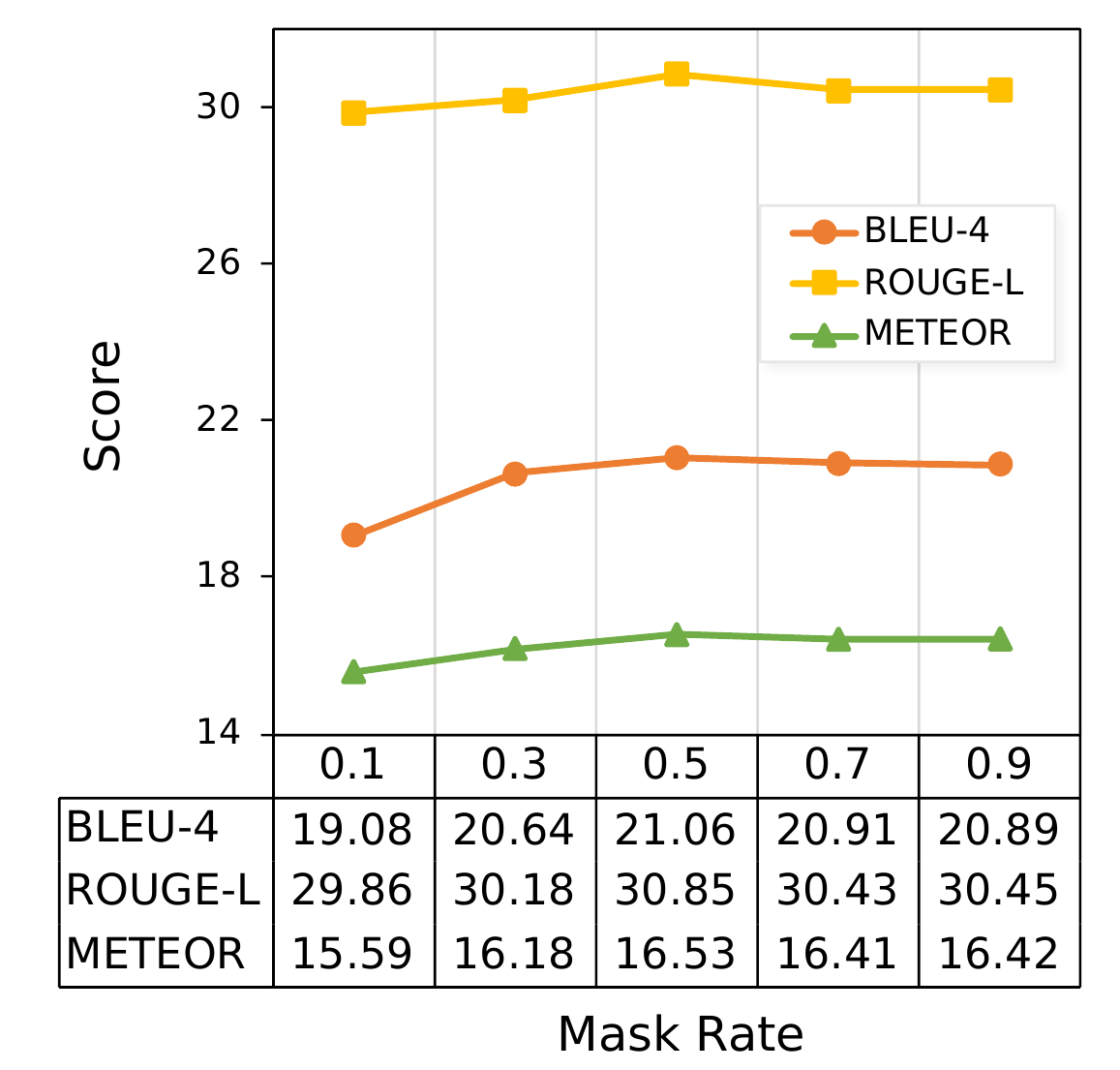}}
\subfloat[Impact of layer number ]{\includegraphics[width=0.3\textwidth]{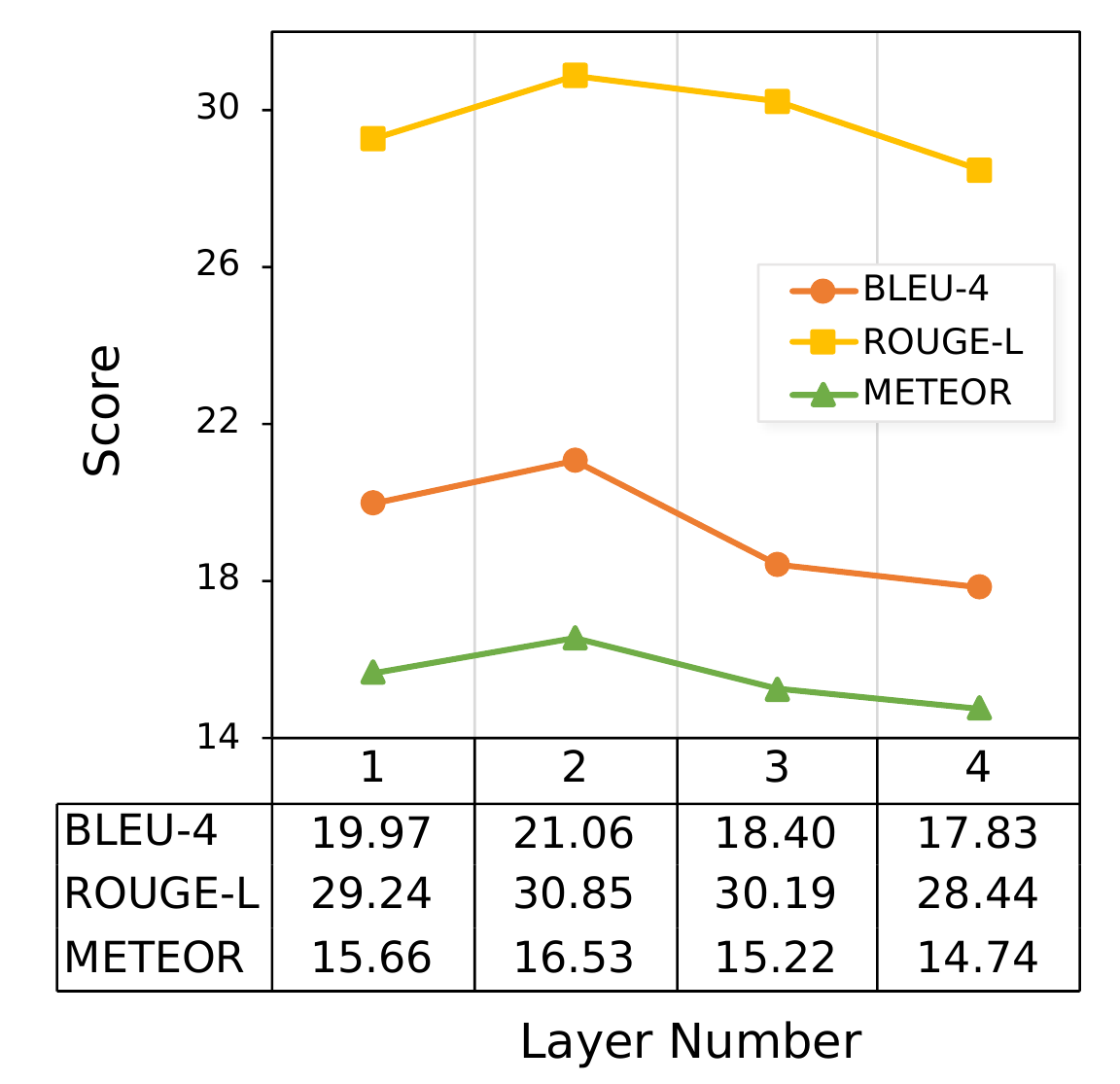}}
\subfloat[Impact of head size]{\includegraphics[width=0.3\textwidth]{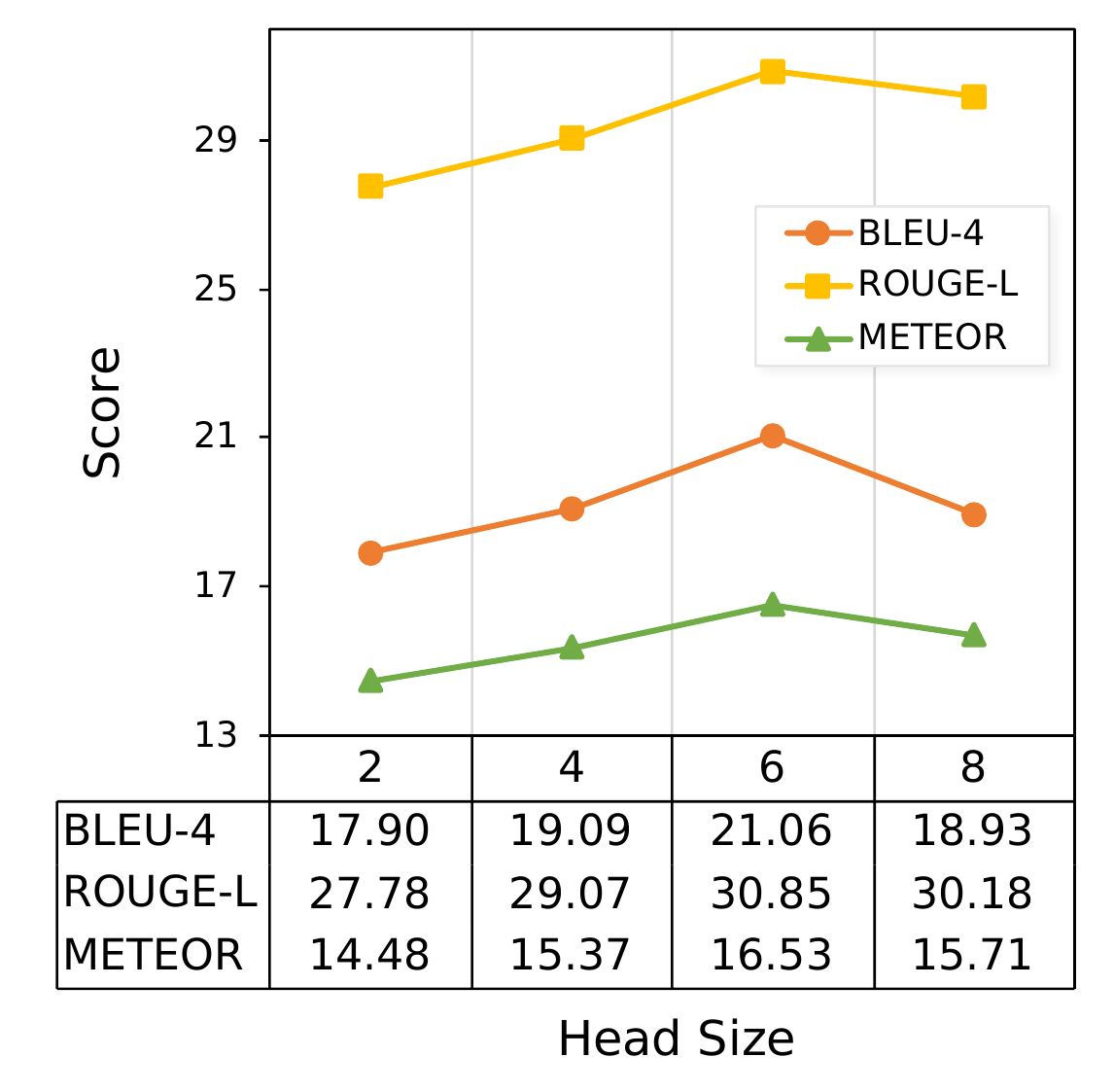}}
\caption{\tool's performance under different hyper-parameter settings. Best results can be achieved by setting the mask rate, layer number and head size to 0.5, 2 and 6, respectively. }
\label{fig:parameter}
\end{figure*}

As we can observe from the comparison in Table \ref{result}, about half of the performance gain compared to the previous state-of-the-art comes from the contextualized code representation learning while the rest can be attributed to the advanced self-attentional model architecture of Transformer. Here, \tool\textsubscript{II}'s substantial improvement compared with the baselines also demonstrates Transformer's strengths over the traditional recurrent-based model architecture in the task domain of code commit message generation. However, a remarkable performance gap still exists between \tool\textsubscript{II} and \tool, which, again, affirms the necessities of contextualized code representation learning in \tool.

\subsection{Parameter Sensitivity}\label{sec:setting}
We further analyze the impact of three key parameters on \tool's performance, including mask rate $\phi$, layer number and head size. Figure \ref{fig:parameter} depicts the analysis results. 

Figure \ref{fig:parameter}(a) shows that the generation quality improves as the mask rate increases from 0.1 to 0.5, but deteriorates as the mask rate keeps increasing. This affirms our hypothesis that masking a continuous fragment \add{can model more code semantics} than masking only a single token, while the adverse impacts of overlarge mask rate can be contrarily explained by the lack of contextual information. In \tool, we set the mask rate to 0.5 in this work, meaning that 50\% tokens of the longest line are randomly masked for Stage I's representation learning. Figure \ref{fig:parameter}(b) and \ref{fig:parameter}(c) implies that, while small layer number or head size reduces performance, excessive number of layers or heads also do harms to model's downstream generation quality. Therefore in \tool, Transformer's layer number and head size are set to 2 and 6 respectively to save computation costs. 

\section{Discussion}\label{sec:discussion}
\subsection{Analysis on the Effects of Data Deduplication}
After a further analysis, we notice that the cleansed dataset released by Liu et al. \cite{liu2018nngen} still contains overlapped code commits across training, validation and test sets. The analysis results are presented in Table \ref{dataset}, where ``Identical Code Changes'' means the commit records containing code changes that are covered in the training set, and ``Completely Identical Entries'' refers to the records having already appeared in the training set with totally same code changes and corresponding commit messages. 

\nie{Since data duplication could adversely affect model performance \cite{allamanis2019adverse}, we then conduct evaluation on the deduplicated dataset, with results shown in Table \ref{supp-bert}. We choose the best retrieval model NNGen \cite{liu2018nngen} and the best generative model NMT \cite{jiang2017automatically} as the baselines. As can be seen, our proposed method \tool still outperforms the baseline models by a significant margin on the deduplicated dataset, which again indicates the efficacy of our proposed contextualized code representation learning framework.}

\begin{table}
 \centering
\caption{Overlapped entries in the benchmark dataset.}
\label{dataset}
\scalebox{0.78}{
\begin{tabular}{|l|c|c|}
\hline
 & \textbf{ Validation Set }   & \textbf{ Test Set} \\ 
\hline
\ Total Entries & 2,511 & 2,521 \\
\ Identical Code Changes & 267 (10.63\%) & 282 (11.19\%) \\
\ Completely Identical Entries & 119 (4.74\%) & 119 (4.72\%) \\ 
\hline
\end{tabular}}
\end{table}

\begin{table}
 \centering
\caption{Evaluation results on the deduplicated dataset.}
\label{supp-dataset}
\scalebox{0.75}{
\begin{tabular}{c|c|c|c|c|c}
\toprule 
\multicolumn{1}{c|}{\multirow{1}{*}{\textbf{Model}}}                                    &  \multirow{1}{*}{\textbf{BLEU-4}}   & \multirow{1}{*}{\textbf{ROUGE-1}}           & 
\multirow{1}{*}{\textbf{ROUGE-2}}  & 
\multirow{1}{*}{\textbf{ROUGE-L}}  & \multirow{1}{*}{\textbf{METEOR}} \\ 
\midrule
\multicolumn{1}{l|}{\ \ NMT } & 10.54 & 20.37 & 10.44 & 19.20 & 9.57  \\ 
\multicolumn{1}{l|}{\ \ NNGen } & 12.44 & 24.22 & 12.04 & 23.76  & 11.66 \\ 
\midrule
\multicolumn{1}{l|}{\ \ \tool} & \textbf{15.86} & \textbf{27.31} & \textbf{15.09} &  \textbf{25.46} & \textbf{13.38}
\\
\bottomrule
\end{tabular}}
\end{table}
\subsection{BERT-Based Approach Comparison}

When trying to comprehend \tool's mechanism, readers may consider the contextualized code representation learning stage (Stage I) as a ``pre-training'' process and the downstream commit message generation stage (Stage II) as a fine-tuning process. However, we adopt the term ``contextualized'' instead of ``pre-trained'' here to distinguish the proposed approach from popular pre-training language models such as GPT \cite{radford2018improving, radford2019language}, BERT \cite{devlin2018bert}, etc. The wording is mainly based on two reasons: 1) popular pre-training language models generally require huge amount of data as training corpus, while only limited size of high-quality data ($\sim$27k pairs of code commits and commit messages) are available in our scenario. 2) popular pre-training language models commonly facilitate multiple downstream tasks \cite{devlin2018bert,feng2020codebert}, while \tool is specifically designed for the commit message generation task. To prevent readers from misunderstanding that we are proposing a general-purpose pre-training approach, we avoid using the terms ``pre-training''/``pre-trained'' in the paper.

To highlight the difference between the proposed \tool and BERT-like pre-trained models, we also compare \tool with the pretrained-BERT-based model \cite{DBLP:conf/iclr/ZhuXWHQZLL20}, named BERT-fused model.
The BERT-fused model also uses Transformer as its base model and fuses the word representations extracted from a pre-trained BERT model with Transformer's encoder and decoder layers. We choose this work as the baseline for two reasons: 1) this work is a representative work that leverages pre-trained BERT model for neural machine translation and achieves state-of-the-art results on several machine translation benchmark datasets; and 2) this work also uses standard Transformer as the basic architecture similar to \tool, therefore eliminating the influence of basic architecture variations.

We use the default hyperparameter settings to implement the BERT-fused model. The experimental results are illustrated in Table \ref{supp-bert}. We can observe that \tool significantly outperforms the BERT-based approach with an increase of 38.92\% in terms of BLEU-4 score. This indicates the effectiveness of \tool's specialized contextualized code representation learning strategies over BERT-based approaches in the task domain of commit message generation.

\begin{table}
 \centering
\caption{Comparison of \tool with BERT-based approach.}
\label{supp-bert}
\scalebox{0.75}{
\begin{tabular}{c|c|c|c|c|c}
\toprule 
\multicolumn{1}{c|}{\multirow{1}{*}{\textbf{Model}}}                                    &  \multirow{1}{*}{\textbf{BLEU-4}}    & \multirow{1}{*}{\textbf{ROUGE-1}}           & 
\multirow{1}{*}{\textbf{ROUGE-2}}           & 
\multirow{1}{*}{\textbf{ROUGE-L}}               & \multirow{1}{*}{\textbf{METEOR}} \\ 
\midrule
\multicolumn{1}{l|}{\ \ BERT-fused} 
& 15.16 & 25.81 & 14.98 & 24.42  & 13.43 
\\ 
\midrule
\multicolumn{1}{l|}{\ \ \tool} & \textbf{21.06} & \textbf{32.87} & \textbf{20.17} &  \textbf{30.85} & \textbf{16.53}
\\
\bottomrule
\end{tabular}}
\end{table}
\subsection{Analysis of \tool with Combined Loss Function }

In \tool, the loss functions of Stage I and Stage II are separated since their respective objectives are essentially different and the model is designed to be optimized in order, i.e., learning the code representations first and then generating commit messages based on the learnt representations. Combining the two losses may bring in undesirable noises along with the task-specific knowledge in training, leading to poor generation results. We conduct a comparison experiment where \tool is associated with a hybrid loss function, named as \tool\textsubscript{Hybrid}. Specifically, during training, with all the other experimental setups kept optimal, Transformer is optimized with a combined loss function:
\begin{equation}
   \mathlarger{\mathcal{L}}(\theta;\mathds{C}) =\mathcal{L}_{\text{I}}+\mathcal{L}_{\text{II}}  ,\end{equation}
where $\mathcal{L}_{\text{I}}$ represents the loss function for Stage I and $\mathcal{L}_\text{II}$ represents the loss function for Stage II, to simultaneously fit in both tasks, i.e. contextualized code representation learning and commit message generation. The experimental results are depicted in Table \ref{supp-hybrid}. As can be seen, the performance of \tool declines dramatically when the two losses are integrated, which indicates the importance of optimizing the model with two separate loss functions sequentially for the task. 

\begin{table}
 \centering
\caption{Comparison of our proposed training method separating the two-stage losses with \tool training with combined loss (denoted as \tool\textsubscript{Hybrid}).}
\label{supp-hybrid}
\scalebox{0.73}{
\begin{tabular}{c|c|c|c|c|c}
\toprule 
\multicolumn{1}{c|}{\multirow{1}{*}{\textbf{Model}}}                                    &  \multirow{1}{*}{\textbf{BLEU-4}}    & \multirow{1}{*}{\textbf{ROUGE-1}}           & 
\multirow{1}{*}{\textbf{ROUGE-2}}           & 
\multirow{1}{*}{\textbf{ROUGE-L}}               & \multirow{1}{*}{\textbf{METEOR}} \\ 
\midrule
\multicolumn{1}{l|}{\ \ \tool\textsubscript{Hybrid}} 
& 15.41 & 22.15 & 11.04 & 20.71  & 13.79 
\\ 
\midrule
\multicolumn{1}{l|}{\ \ \tool} & \textbf{21.06} & \textbf{32.87} & \textbf{20.17} &  \textbf{30.85} & \textbf{16.53}
\\
\bottomrule
\end{tabular}}
\end{table}
\subsection{Analysis of NMT with \tool Framework}

The core idea of contextualized code representation learning in \tool can be flexibly incorporated into other sequence-to-sequence neural architectures. To validate the transferability of our proposed framework, we conduct one supplementary experiment where \tool's Transformer model is replaced with the basic NMT model in Jiang et al.'s work \cite{jiang2017automatically} while keeping the rest experimental setups unchanged. Specifically, the NMT model is first optimized by the objective function as described in Equation 4 in the paper, then get further fine-tuned for downstream commit message generation with Jiang et al.'s default settings. We name this new baseline as NMT\textsubscript{\tool}.

The comparison results are detailed in Table \ref{supp-nmt-1}. As can be observed, by incorporating the contextualized code representation learning framework, NMT\textsubscript{\tool} achieves a significantly better performance over the basic NMT model, presenting an increase of 26.75\% in terms of BLEU-4 score. The result can further exhibit the necessities of contextualized code representation learning in commit message generation, while the performance gap between NMT\textsubscript{\tool} and \tool again demonstrates Transformer's superiority over the traditional NMT model in this task domain.

\begin{table}
 \centering
\caption{Analysis of Jiang et al.'s baseline model (NMT) adopting \tool's two-stage framework (denoted as NMT\textsubscript{\tool}).}
\label{supp-nmt-1}
\scalebox{0.65}{
\begin{tabular}{cc|c|c|c|c|c}
\toprule 
\multicolumn{2}{c|}{\multirow{1}{*}{\textbf{Model}}}                                    &  \multirow{1}{*}{\textbf{BLEU-4}}    & \multirow{1}{*}{\textbf{ROUGE-1}}           & 
\multirow{1}{*}{\textbf{ROUGE-2}}           & 
\multirow{1}{*}{\textbf{ROUGE-L}}               & \multirow{1}{*}{\textbf{METEOR}} \\ 
\midrule
\multicolumn{1}{c|}{\multirow{2}{*}{\textbf{Baselines}}} & \multicolumn{1}{l|}{NMT} 
& 14.17 & 23.12 & 14.36 & 22.09  & 12.54 
\\\multicolumn{1}{c|}{} & \multicolumn{1}{l|}{NMT\textsubscript{\tool}}
& 17.96 & 24.99 & 14.07 & 23.70 & 14.28
\\ 
\midrule
\multicolumn{1}{c|}{\textbf{Ours}}&  \multicolumn{1}{l|}{\tool} & \textbf{21.06} & \textbf{32.87} & \textbf{20.17} &  \textbf{30.85} & \textbf{16.53}
\\
\bottomrule
\end{tabular}}
\end{table}
\vspace{0.1cm}
\begin{table}
 \centering
\caption{\tool's performance with In-statement Code Structure Modeling (ICSM) task integrated.}
\label{supp-future}
\scalebox{0.73}{
\begin{tabular}{c|c|c|c|c|c}
\toprule 
\multicolumn{1}{c|}{\multirow{1}{*}{\textbf{Model}}}                                    &  \multirow{1}{*}{\textbf{BLEU-4}}    & \multirow{1}{*}{\textbf{ROUGE-1}}           & 
\multirow{1}{*}{\textbf{ROUGE-2}}           & 
\multirow{1}{*}{\textbf{ROUGE-L}}               & \multirow{1}{*}{\textbf{METEOR}} \\ 
\midrule
\multicolumn{1}{l|}{\ \ \tool } 
& 21.06 & 32.87 & 20.17 &  30.85 & 16.53
\\ 
\midrule
\multicolumn{1}{l|}{\ \ \tool$_{\text{ICSM}}$} & \textbf{21.37} & \textbf{32.88} & \textbf{20.33} &  \textbf{32.87} & \textbf{16.72} \\
\bottomrule
\end{tabular}}
\end{table}

\subsection{Future Research Direction}
According to the task nature, various representation learning methods can be also integrated together to maximize the exploitation and utilization of code's contextual and structural information. In \tool, we propose \textit{Code Changes Prediction} and \textit{Masked Code Fragment Prediction} tasks to model the code contextual information corresponding to the two separate categories of code commits. In future, these tasks can be further extended to include more complicated and well-designed representation learning methodologies. 

For example, comparing to natural language, the syntactic structure of code are more rigid. Tokens in the same code statement are generally of stronger semantic relations than the tokens from other statements. Therefore, we design an additional code representation learning task to model this in-statement code structural information. 

Specifically, inspired by the idea of pairwise code encoding \cite{ahmad2020transformer}, in Stage I, we additionally train the model to predict a randomly masked code token from the other tokens in the same code statement. Formally, given a source code sequence $X$ that can be split into a set of $n$ code statements $\{X^1, X^2, ..., X^n\}$, for each statement $X^i$, we randomly mask a token $x^i_u \in X^i$, then predict this masked token based on the remaining tokens from the same statement $\{x^i_1, ..., x^i_{u-1}, x^i_{u+1}, ..., x^i_{m}\}$ using the log likelihood objective function:
\begin{equation}
    \mathlarger{\mathcal{L}}_{3}=-
    \sum_{X\in \mathds{C}_1} \sum_{X^i\in X} \log \mathcal{P}(x^{i}_{u}|x^i_1, ..., x^i_{u-1}, x^i_{u+1}, ..., x^i_{m};\theta), 
\end{equation}
where $\mathds{C}_1$ refers to the same commit subcorpora with explicit code changes as in Equation (1). Thereby, the attention inside Transformer can be subtly guided to flow among the code tokens of the same statement, allowing the model to capture the code structural information more effectively and achieve more accurate contextualized code representation learning. 

The experimental results with the above method integrated are shown in Table \ref{supp-future}. As can be seen, the in-statement code structure modeling task further boosts \tool's performance on downstream commit message generation. This promising result suggests the great potentials of \tool in more effectively exploiting code contextual and structural information with other representation learning tasks integrated. In future, we will consider embedding code structural graphs, such as control flow graph and program dependency graph, for a more accurate modeling of the code contextual information.
\section{Related Works}\label{sec:literature}
This section reviews the most related works and groups them into three lines: commit message generation, contextualized word representation, and code representation learning.

\subsection{Commit Message Generation}
Existing literature for commit message generation can be roughly divided by their methodologies into three categories: rule-based, retrieval-based and deep-learning-based. 

Earliest works in the field attempt to automate the commit message generation by extracting information from the code commits and filling in pre-defined templates \cite{buse2010automatically, cortes2014automatically, linares2015changescribe, shen2016automatic}. Among them, Shen et al. \cite{shen2016automatic} use pre-defined formats to identify the commit type and generate commit messages based on corresponding templates. ChangeScribe \cite{linares2015changescribe} further takes the impact set of a commit into account when extracting core information from the code commits. In spite of the involvement of prior knowledge, these rule-based methods can only handle the code commits that match certain formats and the produced commit messages can only cover trivial commits.

Therefore, later works leverage informational retrieval techniques to allow more flexible commit message generation \cite{huang2017mining, liu2018nngen}. For example, Huang et al. \cite{huang2017mining} evaluate the similarity among code commits based on both syntactic and semantic analysis and reuse the message of the most similar commit as model output. NNGen \cite{liu2018nngen} generalizes the similarity measurement by calculating the cosine distance between bag-of-words vectors of the code commits, which extends to also support the code commits with implicit binary file changes. However, retrieval-based approaches are still limited in two aspects: the variable/function names are usually not consistent in the retrieved message, and the generation performance relies heavily on the coverage of the database. 

By adopting deep neural networks to translate code commits into messages, deep-learning-based methods have gradually become the mainstream approach in this research field. Both Loyola et al. \cite{loyola2017neural} and Jiang et al. \cite{jiang2017automatically} propose to bridge the gap between code commits and commit messages with an attentional encoder-decoder framework. Loyola et al.'s later work \cite{loyola2018content} further takes intra-code documentation as a guiding element to improve the generation quality. Since deep learning models suffer heavily from context-specific tokens, CODISUM \cite{xu2019commit} and PtrGNCMsg \cite{liu2019generating} both attempt to mitigate OOV issue by incorporating the copying mechanism, while the former one fails in supporting the code commits with implicit binary file changes. In all these methods, code contextual information is either neglected or built up using an additional network.

\subsection{Contextualized Word Representation}
Our work also relates closely to the contextualized word representation methods. Pioneering word representation methods keep the mapping function invariant across different sentences \cite{mikolov2013distributed, pennington2014glove, bojanowski2017enriching}. For example, Word2vec learns the word embedding by a skip-gram or continuous-bag-of-word (CBOW) model, which are both based on distributed center-context word pair information \cite{mikolov2013distributed}. Comparatively, Glove produces word embeddings by factorizing the word co-occurrence matrix to leverage global statistical information contained in a document \cite{pennington2014glove}. Although these methods can capture both syntactic and semantic meanings behind the words, the limitation of these static word embedding approaches lies mainly in two aspects: 1) these approaches do not leverage the information of entire sentence and the relationships learned from the center-context pairs are restricted in fixed window-size, and 2) these approaches fail to capture polysemy since the embedding tables are kept invariant across different contexts.

In recent years, contextualized word representation methods have gained overwhelming dominance. Pre-trained from large unlabeled corpus, contextualized word representations can capture word sense, syntax, semantic roles and other information dynamically from the context, achieving state-of-the-art results on many downstream tasks including question answering, sentiment analysis, reading comprehension, etc \cite{peters2018deep, radford2018improving, radford2019language, devlin2018bert}. Specifically, Peter et al. \cite{peters2018deep} derive the word representations from a bi-directional LSTM trained with coupled language model objective on a large corpus. The GPT model proposed by OpenAI instead uses multi-layer Transformer decoders for the language model pre-training \cite{radford2018improving, radford2019language}. However, the left-to-right architecture of GPT models can be harmful for many token-level tasks where the contextual information from both directions are equally essential. Therefore, to alleviate the unidirectional nature of language models, Devlin et al. \cite{devlin2018bert} further pre-train a denoising auto-encoder using a brand new self-supervised learning task named ``masked language model''. By predicting the randomly masked word tokens from their contexts, contextualized word representations are embedded into the initialized model parameters for downstream tasks' usage. Unlike static word representation methods that require an extra network for downstream task processing, these networks can be adapted to various downstream tasks with simple architecture modifications.

\subsection{Code Representation Learning}
Among the previous works of code representation learning, traditional machine learning algorithms used to be the standard practices. In particular, by treating the code as a sequence of tokens, n-gram language model was widely adopted in modeling the source code for authorship classification \cite{frantzeskou2008examining}, repository mining \cite{allamanis2013mining}, convention detection \cite{allamanis2014learning}, etc. SVM is another common approach for representing the programs that has been applied for malicious code detection \cite{choi2011efficient} and code domain categorization \cite{linares2014using}. By further taking the syntax tree structure of code into consideration, Maddison \& Tarlow \cite{maddison2014structured} describe new generative models based on probabilistic context-free grammars, while Raychev et al. \cite{raychev2016probabilistic} build up code probabilistic model by learning decision trees out of a domain-specific language called TGen. 

Recent advancement of deep learning models also changes the way researchers representing code semantics. Token-based techniques process code as textual data and adopt RNN models to learn the code features together with downstream tasks \cite{raychev2014code}. Tree-based techniques transform syntax tree into vectors that are later formatted as model input. For example, for code defect prediction, Wang et al. \cite{wang2016automatically} leverage a deep belief network to learn the semantic code representations from abstract syntax tree (AST) nodes, while for code clone detection, White et al. \cite{white2016deep} use a recursive auto-encoder to exploit code syntactical information from ASTs. TBCNN \cite{mou2016convolutional} includes a tree-based convolution on ASTs to learn program vector representations. ASTNN \cite{zhang2019novel} decomposes large ASTs into sequences of small statement trees and finally learns the code representation from encoded statement vectors. Last category of graph-based techniques constructs the entire syntax graph as model input. Allamanis et al. \cite{allamanis2017learning} leverage a Gated Graph Neural Network to represent both the syntactic and semantic structure of source code. Compared with these methods, our proposed approach focuses on learning contextualized code representation without using external ASTs or constructed graphs, which can achieve a greater balance between the performance and usability of downstream commit message generation. 

Inspired by the success of the aforementioned pre-trained language models, SCELMo \cite{karampatsis2020scelmo} and CodeBERT \cite{feng2020codebert} propose to pre-train code representation on large unlabeled corpus. However, these works directly borrow the pre-training tasks from original implementations without explicitly taking into account the semantic gaps between source code and natural language.
\section{Conclusion}\label{sec:con} 
Code commit message generation is a necessitated yet challenging task. In this paper, we proposed \tool, a two-stage framework that takes advantage of contextualized code representation learning to boost the downstream performance of commit message generation. Specifically, with regard to the two categories of code commits, we introduce two representation learning strategies, namely \textit{Code Changes Prediction} and \textit{Masked Code Fragment Prediction}, for the exploitation of code contextual information. Experimental results showed that \tool significantly outperforms competitive baselines and achieves the state-of-the-art on the benchmark dataset. 

\add{\tool is also validated under low-resource settings, where high quality commit messages were generated with only 50\% of the labels utilized during the fine-tuning. This points out promising future directions of extending this contextualized code representation learning framework to larger code corpus and other similar code-related tasks, such as code summarization. Moreover, \tool's improvements after exploiting the in-statement code structure also demonstrate its great potentials in integrating more complicated code contextual and structural information in future. }



\section{Acknowledgement}
This work was supported by the National Natural Science Foundation of China under project No. 62002084, and partially supported by a key program of fundamental research from  Shenzhen Science and Technology Innovation Commission (No. ZX20210035), Singapore Ministry of Education Academic Research Fund Tier 1 (Award No. 2018-T1-002-069), the National Research Foundation, Prime Ministers Office, Singapore under its National Cybersecurity R\&D Program (Award No. NRF2018NCR-NCR005-0001), the Singapore National Research Foundation under NCR Award Number NRF2018NCR-NSOE003-0001, NRF Investigatorship NRFI06-2020-0022.

\bibliographystyle{cas-model2-names}

\bibliography{ref.bib}



\end{document}